\documentclass[accepted]{uai2021} % after acceptance, for a revised
\usepackage[american]{babel}
\usepackage{natbib} % has a nice set of citation styles and commands
\usepackage{mathtools} % amsmath with fixes and additions
\usepackage{booktabs} % commands to create good-looking tables
\usepackage{tikz} % nice language for creating drawings and diagrams
\usepackage{amsfonts}
\title{RECOWNs: Probabilistic Circuits for Trustworthy Time Series Forecasting}

\author[1]{\href{mailto:Nils Thoma <nthoma@nilsthoma.de>}{Nils~Thoma}{}}
\author[1]{Zhongjie Yu}
\author[1]{Fabrizio Ventola}
\author[1,2]{Kristian Kersting}
\affil[1]{%
    Department of Computer Science, TU Darmstadt, Darmstadt, Germany
}
\affil[2]{%
    Centre for Cognitive Science, TU Darmstadt, and Hessian Center for AI (hessian.AI)
}

\begin{document}
\maketitle

\begin{abstract}
    Time series forecasting is a relevant task that is performed in several real-world scenarios such as product sales analysis and prediction of energy demand. 
    Given their accuracy performance, currently, Recurrent Neural Networks (RNNs) are the models of choice for this task. 
    Despite their success in time series forecasting, less attention has been paid to make the RNNs trustworthy. 
    For example, RNNs can not naturally provide an uncertainty measure to their predictions.
    This could be extremely useful in practice in several cases e.g. to detect when a prediction might be completely wrong due to an unusual pattern in the time series.
    Whittle Sum-Product Networks (WSPNs), prominent deep tractable probabilistic circuits (PCs) for time series, can assist an RNN with providing meaningful probabilities as uncertainty measure.
    With this aim, we propose RECOWN, a novel architecture that employs RNNs and a discriminant variant of WSPNs called Conditional WSPNs (CWSPNs). %
    We also formulate a Log-Likelihood Ratio Score as an estimation of uncertainty that is tailored to
    time series and Whittle likelihoods.
    In our experiments, we show that RECOWNs are accurate and trustworthy time series predictors, able to
    ``know when they do not know''.
\end{abstract}

\section{Introduction}
\label{sec:intro}

        \begin{figure}[bt!]
            \graphicspath{{./plots/}}
              \centering
              \includegraphics[width=0.9\linewidth]{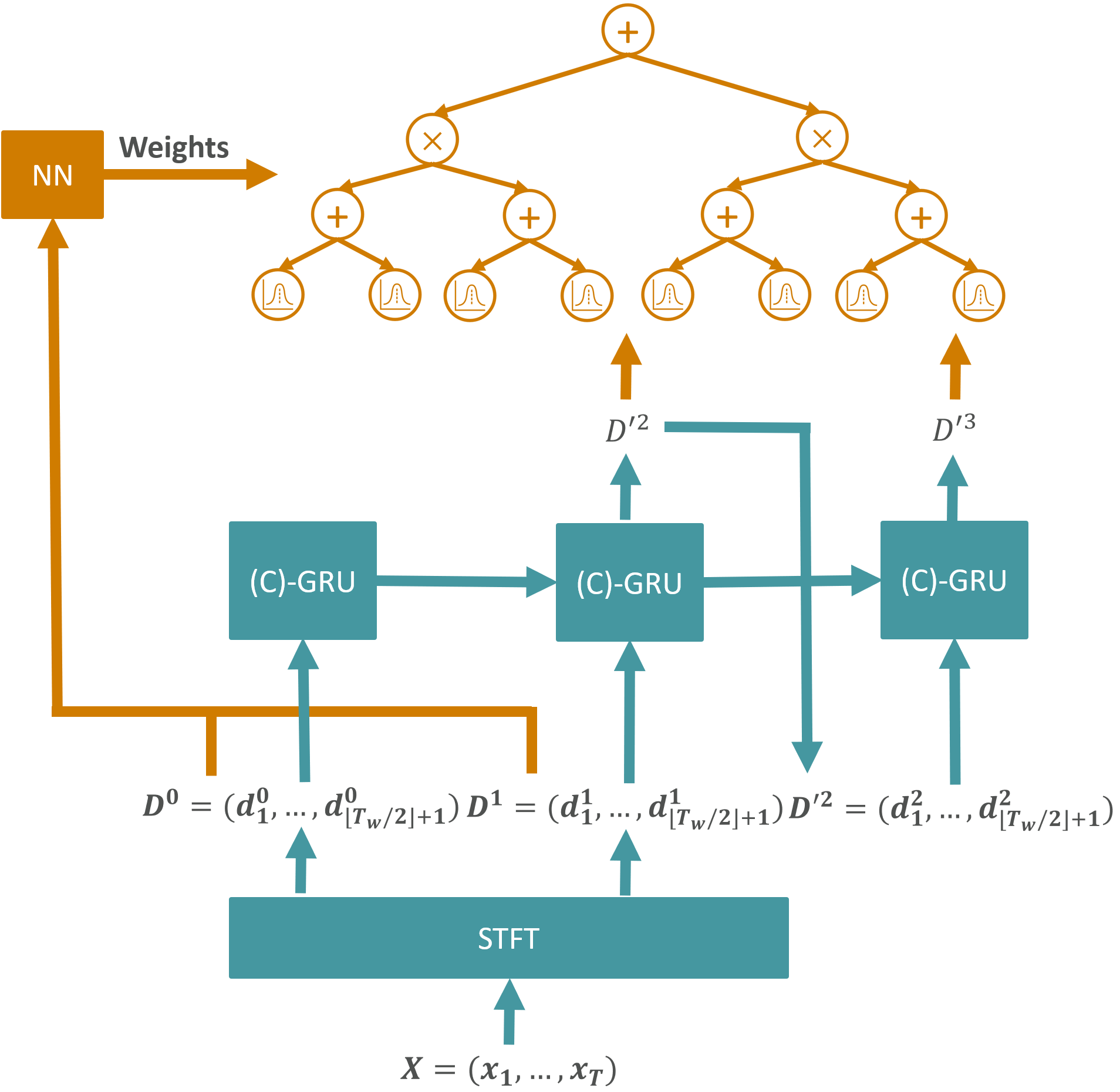}
            \caption{Overview of the RECOWN architecture. The context $X$ is transformed using STFT with a window size of $T_w$, a) determining the weights of the CWSPN via the Neural Network (NN) and b) serving as input to the (complex)-Gated Recurrent Units (GRU)~\citep{chung2014empirical} of the RNN, resulting in the prediction of the Fourier coefficients $D'^2, D'^3$. Those are then provided to the CWSPN, which computes the conditional Whittle log-likelihood $\ell(D'^2, D'^3 | D^0, D^1)$ (see Section \ref{sec:cCSPN}).}
            \label{fig:SystemStructure}
        \end{figure}

        \begin{figure*}[t!]
            \graphicspath{{./plots/}}
              \centering
              \includegraphics[width=.89\textwidth]{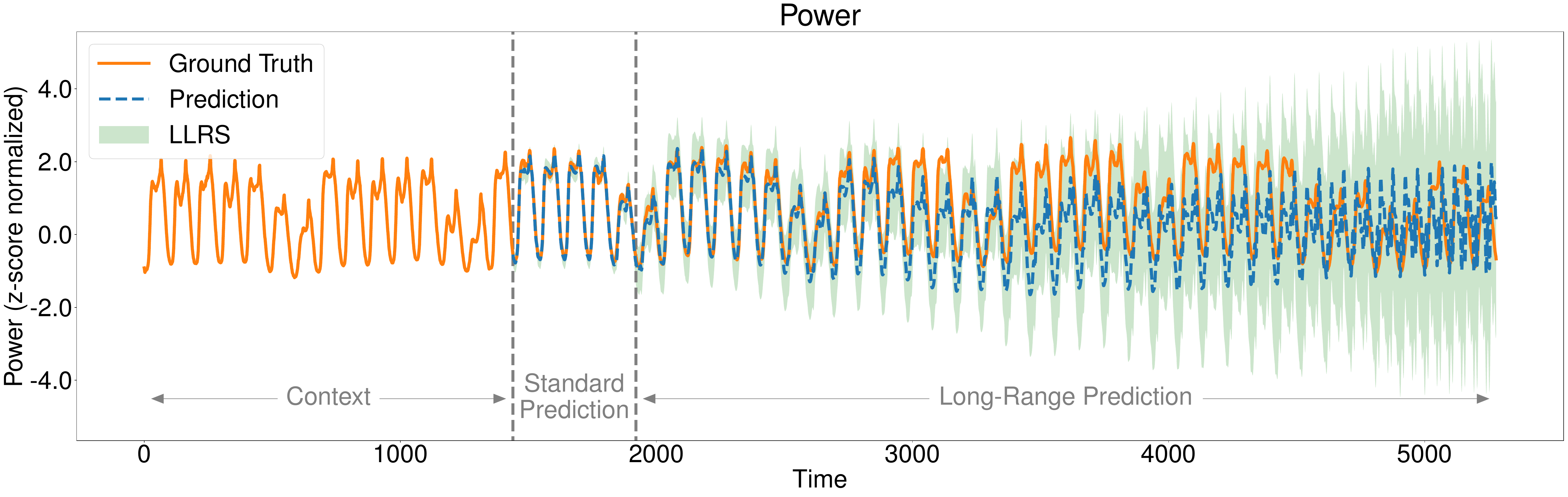}
              \includegraphics[width=.89\textwidth]{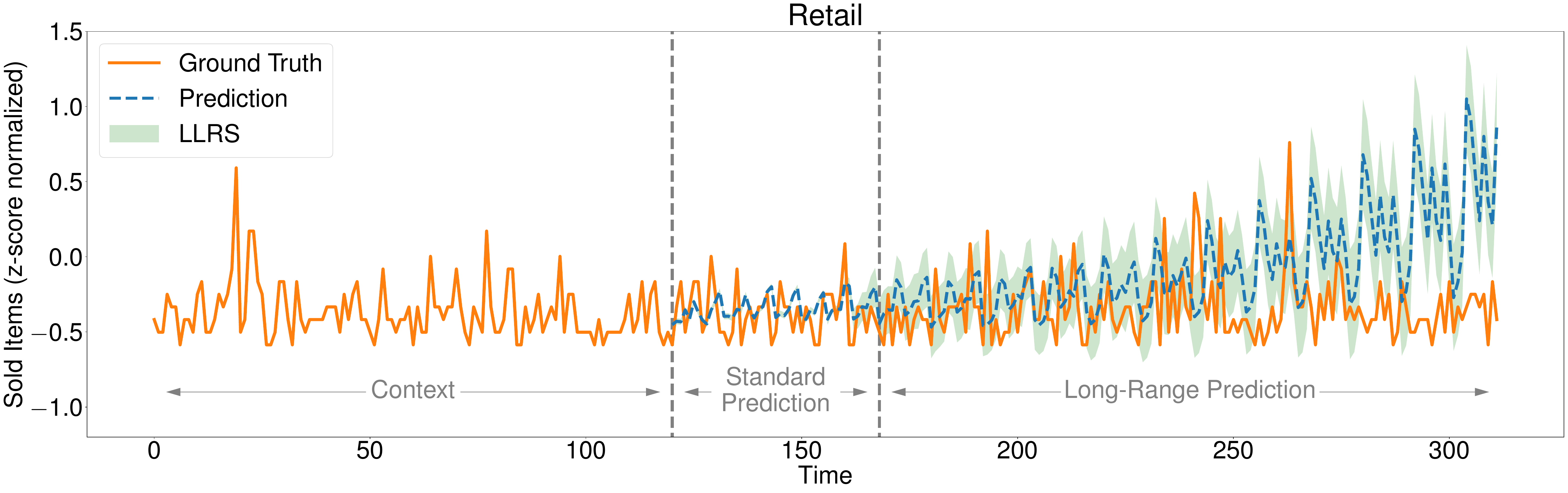}
             \caption{A long-range prediction on \textit{Power} (top) and \textit{Retail} datasets (bottom). RECOWNs are trustworthy models that can provide predictions with an uncertainty measure estimated by the Log-Likelihood Ratio Score (LLRS, in green, see Section \ref{sec:Uncertainty}). This can be provided at any time point of the prediction. Here, the model has been trained to perform short-range ``standard'' forecast. Obviously, the model is less accurate
             on longer-range predictions. For those, RECOWNs correctly indicate their confidence with a meaningful increase of uncertainty (LLRS) over time.
             This makes the model trustworthy and can support users in decision-making processes.}
             
            \label{fig:Uncertainty}
        \end{figure*}

    Time series forecasting is the task to predict the future course ($Y$) of a time series given its past $X$, also known as context.
    Currently, Recurrent Neural Networks~\citep{rumelhart1985learning} are models of choice when it comes to time series forecasting. 
    Recent advancements in RNNs research fostered their adoption in practice surpassing established models, e.g. ARIMA (Autoregressive Integrated Moving Average)~\citep{siami2018comparison}, in several scenarios~\citep{fei2015temporal, li2019convolutional, li2018prediction}. 
    However, in complex real-world applications, time series are highly subject to several influence factors
    which are often hard to capture.
    For example, in the case of grocery demand, the demand for ice cream or BBQ-related products is expected to be higher as long as the weather is warmer than usual. 
    Furthermore, exceptional events like a pandemic can significantly influence demands as well.
    In such cases, the prediction of a model will likely be less accurate compared to usual circumstances. 
    To properly detect such cases, a measure of uncertainty about the prediction is valuable~\citep{guo2017calibration, laptev2017time, gal2016dropout}. It would make the predictions trustworthy and it would better support the users in decision-making processes.
    A relevant approach to achieve this is provided by Gaussian Processes (GPs)~\citep{RasmussenW06}.
    However, GPs are comparably computationally expensive and therefore not suited for large datasets~\citep{seeger2004gaussian}.
    Several methods have been proposed to scale GPs on large datasets, e.g. by modeling the mixture of subspaces using GP experts with Sum-Product Networks (SPNs)~\citep{trapp2020deep, BruinsmaPTHST20, yu2021uai_momogps}. 
    Anyhow, for time series forecasting, RNN architectures~\citep{alpay2016learning, wolter2020sequence, koutnik2014clockwork} have shown to be superior in terms of prediction accuracy.
    \citet{pmlr-v139-rasul21a} already examined this direction, by employing auto-regressive denoising diffusion models in the time domain to equip RNN-predictions with a measure of uncertainty on the level of time steps.
    Still, they do not provide an uncertainty measure on the level of sequences, which enables users to quickly detect potentially problematic forecasts.
    Furthermore, modeling in the spectral domain is beneficial for prediction performance~\citep{wolter2020sequence}.
    
    Therefore, we introduce REcurrent COnditional Whittle Networks (RECOWNs), 
    a deep architecture that makes use of \nameref{sec:STFT} (STFT) and
    integrates a Spectral RNN~\citep{wolter2020sequence} together with a conditional variant of a Whittle SPN~\citep{yu2021icml_wspn} (CWSPN) i.e.
    a discriminative probabilistic circuit tailored for time series. 
    The overall architecture is depicted in Figure~\ref{fig:SystemStructure}.
    In this way, RECOWN can keep the good forecasting performance of RNNs while, thanks to the CWSPN, providing meaningful probabilities as a measure of uncertainty of the predictions.
    A showcase of its potential can be found in Figure~\ref{fig:Uncertainty}, showing how the CWSPN correctly predicts
    when the forecast of the Spectral RNN increasingly diverges from the ground truth.
    RECOWNs can be trained end-to-end by gradient descent and have modular nature, in fact, one can employ
    any differentiable density estimator instead of a CWSPN. 
    However, in our experiments, CWSPNs resulted in the best choice compared to another state-of-the-art density estimator, MAF~\citep{papamakarios2017masked}, especially when the model capacity is reduced.
    Furthermore, CWSPNs are also more flexible given that they can answer to a wider range of exact
    probabilistic queries in a tractable way.
    
    Similar to~\citet{yu2021icml_wspn}, RECOWNs model time series in the spectral domain, but employ a
    different kind of Fourier Transform -- the STFT -- to account for changes in the frequency domain as the signal changes over time.  
    While ~\citet{yu2021icml_wspn}
    focus on modeling the joint distribution of the multivariate time series, in this work we shift our
    focus on predictions, providing them with probabilities as a useful measure of uncertainty.
    As successfully done in~\citet{yu2021icml_wspn} and other previous methods~\citep{tank2015bayesian}, modeling time series in the spectral domain enables us to make use of the Whittle Approximation Assumption \citep{whittle1953analysis} which facilitates the modeling of Fourier coefficients of a time series. 
    Additionally, since RECOWNs model the time series in the spectral domain, we propose Log-Likelihood Ratio Score (LLRS) which enables us to compute the confidence intervals of the predictions back in the time domain.
    Experiments show that, compared to state-of-the-art models, RECOWNs are more accurate and trustworthy. Our contributions are the following:
    \begin{itemize}
        \item We introduce RECOWN, the first deep tractable model for time series forecasting that is
        accurate and that provides an uncertainty measure for its predictions on sequence level as well as in the time domain, making them more trustworthy.
        \item We introduce a data sample weighting strategy based on the Mean Squared Error.
        \item We formulate the (Whittle) Log-Likelihood Ratio Score, tailored to better estimate the uncertainty of time series predictions based on the Whittle likelihood.
    \end{itemize}
    
    The paper is structured as follows: We start by introducing STFT, CWSPNs as well as Spectral RNNs, i.e.
    the main components of RECOWNs in Section \ref{sec:CWSPNs}. Then, in Section \ref{sec:Experiments}, we
    describe the experimental setting and analyze the results, showing that RECOWNs are accurate as well as
    trustworthy and that the LLRS is a valid tool to assess the uncertainty and the quality
    of the predictions.
    We conclude in Section \ref{sec:Conclusion} where we also point out future directions.

\section{RECOWN: Recurrent Conditional Whittle Networks}
\label{sec:CWSPNs}

    \subsection{Short Time Fourier Transform}
    \label{sec:STFT}
    
        With Discrete Fourier Transformation (DFT), a time series can be mapped from the time to the spectral domain with a decomposition into linear combinations of sinus functions.  
        For a multivariate time series $\mathcal{X} = x_1, ..., x_{T}$ with $x_t \in \mathbb{R}^p$ and length $T$, we can define the discrete Fourier coefficients $d_k \in \mathbb{C}^p$ at frequency $\lambda_k = \frac{2 \pi k}{T}$, $k=0, \cdots , T-1$, using the Fourier transformation $\mathcal{F}$ as follows \citep{tank2015bayesian}:
		\begin{equation}
		\label{eq:FT}
			\mathcal{F}(\mathcal{X})_k = d_k = \sum_{t = 0}^{T-1} x_t e^{-i \lambda_k t}.
		\end{equation}
		Moreover, given Fourier coefficients $D = (d_0, ..., d_{T-1})$, we can apply the inverse DFT to project the frequencies back to the time domain:
		\begin{equation}
		\label{eq:FTInv}
			\mathcal{F}^{-1}(D)_t = x_t = \frac{1}{T}\sum_{k = 0}^{T-1} d_k e^{i \lambda_k t}.
		\end{equation}
        
        However, to account for potential changes in the frequency domain as the signal changes over time, we apply STFT.
        The input is divided into overlapping segments and each segment is then approximated separately.
        STFT is introduced by \citet{griffin1984signal} by considering segments of length $T_w$, extracted every S time steps:
		\begin{equation}
		    \begin{aligned}
			    STFT(\mathcal{X})_k^m & = \mathcal{F}(w(Sm - t) x_t)_k \\
			    & = \sum_{t = 1}^{T_w} w(Sm - t) x_t e^{-i \lambda_k t},
		    \end{aligned}
		\end{equation}
		with $w$ being a window function, $m$ denoting the corresponding shift and the remaining defined analogously to Equation~\ref{eq:FT}.
		This results in $n_s = \frac{T - T_w}{S} + 3$ windows in total when applying a padding of size $S$ at the start and end of sequences.
		As with the regular Fourier transform, STFT can also be inverted:
		\begin{equation}
            \begin{aligned}
			iSTFT(D)_t & = \mathcal{F}^{-1}(D(Sm))_t = x_t \\
			& = \frac{\sum_{m = -\infty}^{\infty} w(Sm - t) \mathcal{F}^{-1}(D(Sm)_t)}{\sum_{m = -\infty}^{\infty} w^2(Sm - t)}.
            \end{aligned}
		\end{equation}
		Due to Hermitian symmetry, for real-valued time series, the negative Fourier coefficients are redundant.
		Therefore, we only need to model $\mathcal{T} = \left \lfloor \frac{T_w}{2} \right \rfloor + 1$ Fourier coefficients for a window size of $T_w$.
		Furthermore, we apply a low-pass filter to filter out noise, which further reduces the number of parameters in the model.
		Details on the extent of filtering are described in Section \ref{sec:Experiments}.
		
		The window function can be determined either by hand or learned by an optimizer \citep{wolter2020sequence}.
		We apply a truncated Gaussian window at position $n$:
		\begin{equation}
		    w(n) = \exp \left(-\frac{1}{2} \left(\frac{n - T_w / 2}{\sigma T / 2}\right)^2\right).
		\end{equation}
		The standard deviation $\sigma$ is learned by the optimizer.
		By increasing $\sigma$, the window approaches a more rectangular shape, while it gets narrower when decreasing $\sigma$.

    \subsection{Whittle Likelihood}
    \label{sec:Whittle}
    
        \begin{figure*}[t!]
            \graphicspath{{./plots/}}
             \centering
            \includegraphics[width=.7\textwidth]{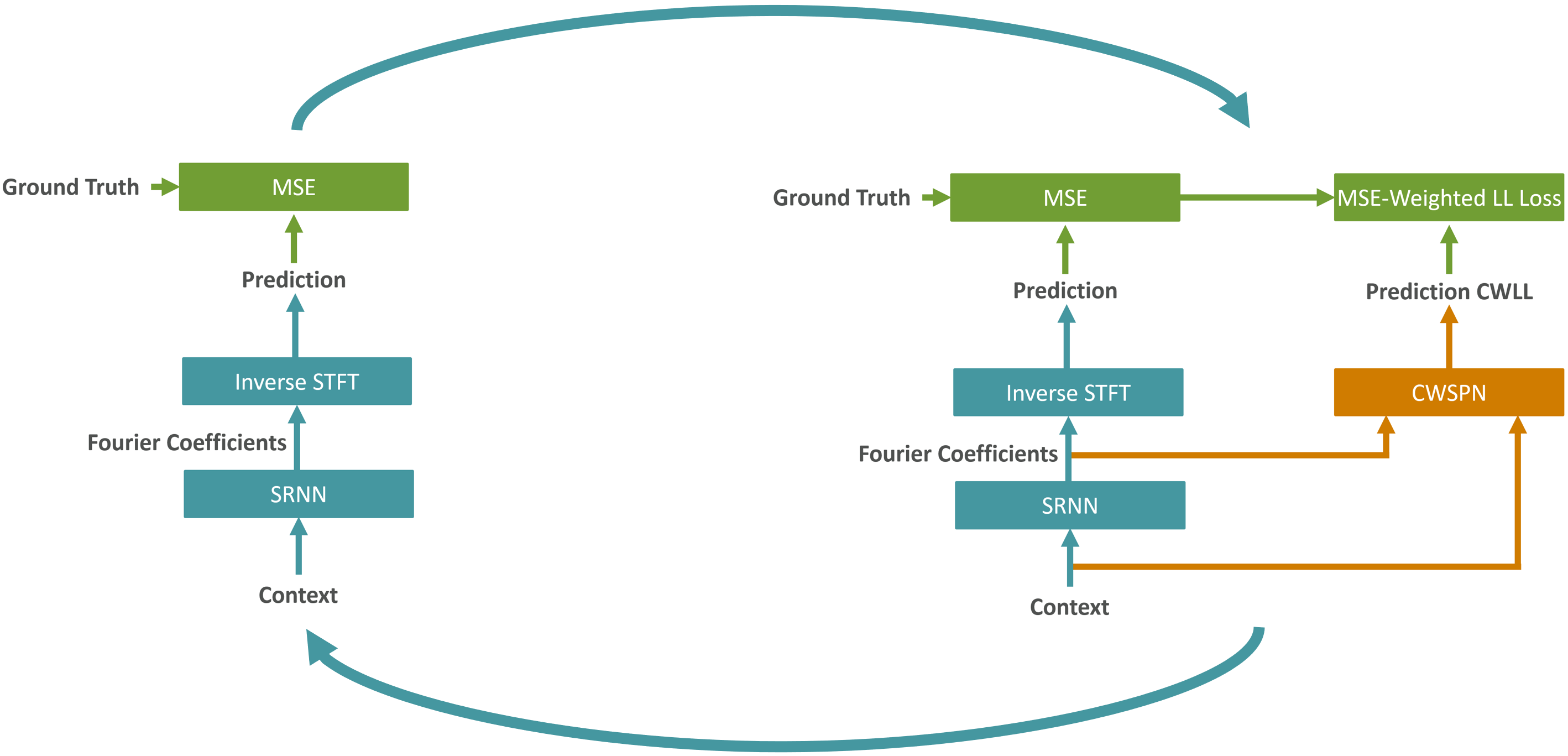}
            \caption{Depiction of the optimization process for RECOWN in co-ordinate descent fashion. First, SRNN weights are optimized (Left), then CWSPN weights are optimized using the MSE together with the CWLL, which is derived from the SRNN predictions (Right). These steps are iterated until convergence.}
            \label{fig:Training}
        \end{figure*}
    
        The Whittle likelihood models multivariate time series in the spectral domain.
        Given $\mathcal{X}$, $x_t$ and $T$ as defined in Section \ref{sec:STFT}, $x_t$ is Gaussian stationary for $t \in \mathbb{Z}$, if:
        \begin{align}
        \label{eq:W1}
        E(x_t) &= \mu  \quad \forall t \in \mathbb{Z} \\
        \text{Cov} (x_t, x_{t+h}) &= \Gamma(h) \quad \forall t,h \in \mathbb{Z}.
        \end{align}
        Given $\mathcal{X}_{1:N} = \{ \mathcal{X}^1, \ldots , \mathcal{X}^N \}$ being $N$ independent realizations of $\mathcal{X}$ and $d_{n, k} \in \mathbb{C}^p$ represent the discrete Fourier coefficient of the $n$-th
        sequence at frequency $\lambda_k$:
        \begin{equation}
        d_{n, k} = \mathcal{F}(\mathcal{X}^n)_k.
        \end{equation}
        Based on the Whittle approximation assumption~\citep{whittle1953analysis}, the Fourier coefficients are independent complex normal RVs with mean zero:
        \begin{equation}
        d_{n, k} \sim \mathcal N (0, S_k), \quad k=0, \ldots , T-1,
        \label{eq:complex_normal}
        \end{equation}
        with $S_k \in \mathbb{C}^{p \times p}$ being the \textit{spectral density matrix}. 
        For a stationary time series, it is defined as:
        \begin{equation}
         S_k = \sum\nolimits _{h=-\infty}^{\infty} \Gamma (h) e^{-i \lambda_k h},
        \label{eq:spectral_density_matrix}
        \end{equation}
        where the infinite sum may be approximated.
        Finally, the Whittle likelihood of the $\mathcal{X}_{1:N}$ is given by: $p(\mathcal{X}_{1:N} \mid S_{0:T-1}) \approx$
        \begin{equation}
         \prod\nolimits_{n=1}^{N} \prod\nolimits_{k=0}^{T-1} \frac{1}{\pi ^p \left | S_{k} \right |} e^{-d_{n, k}^{*}S_{k}^{-1} d_{n, k}}.
        \label{eq:whitle_likelihood_0}
        \end{equation}
        The Whittle approximation holds asymptotically with large $T$ and has been used also in Bayesian context \citep{tank2015bayesian}.
        We will make use of it and place a \nameref{sec:cCSPN} over the frequencies, resulting in CWSPNs, modeling the conditional Whittle Log-Likelihood (CWLL).
    
    \subsection{Complex Conditional SPN}
    \label{sec:cCSPN}
        
        Sum-Product Networks \citep{poon2011sum} are deep probabilistic circuits with tractable and exact inference. Among several applications, they have been successfully employed
        also for univariate time series modeling \citep{melibari2016dynamic}.
        Recent tensor-based SPN implementations such as RAT-SPNs \citep{peharz2020random} build upon randomly generated structures based on the notion of region graphs \citep{dennis2012learning}. They are more scalable and enable training in an end-to-end fashion which allows them to be optimized jointly with NNs.%, making it a perfect fit for our approach.
    
        To provide a measure of how good a prediction ($Y$) is with respect to a context ($X$), we aim for modeling the conditional likelihood $P(Y|X)$.
        Although any SPN modeling the joint distribution could be employed for this task (since the conditional can be derived from the joint using $P(Y | X) = \frac{P(X, Y)}{P(X)}$ ), the Conditional SPN (CSPN)~\citep{shao2020conditional} is a more natural choice.
        CSPN parameters are not learned directly,
        instead, a separate general function approximator $g$ -- in our case a neural network -- is employed to provide parameters $\theta$ to the SPN based on the input x, i.e. $g(x) = \theta$, while only $y$ is provided as input to the SPN on the leaf layer.
        Analogously to \citet{shao2020conditional}, we structure $g$ as two separate, fully-connected networks $g_L$ and $g_S$ with ReLU activations, learning the leaf and sum node parameters $\theta = (\theta_L$, $\theta_S)$ respectively, i.e. $g(x) = (g_L(x), g_S(x))$.
        Training this architecture -- using gradient descent -- results in modeling $P(Y=y|X=x) := P(Y=y;\theta)$.
        The structure of CSPN is generated at random, based on RAT-SPNs~\citep{peharz2020random}.
        In this way, we can take advantage of the benefits provided by a tensorized SPN implementation.
        
        \begin{figure*}[t!]
            \graphicspath{{./plots/}}
            \centering
            \begin{minipage}[b]{.495\textwidth}
              \centering
              \includegraphics[width=0.9\textwidth]{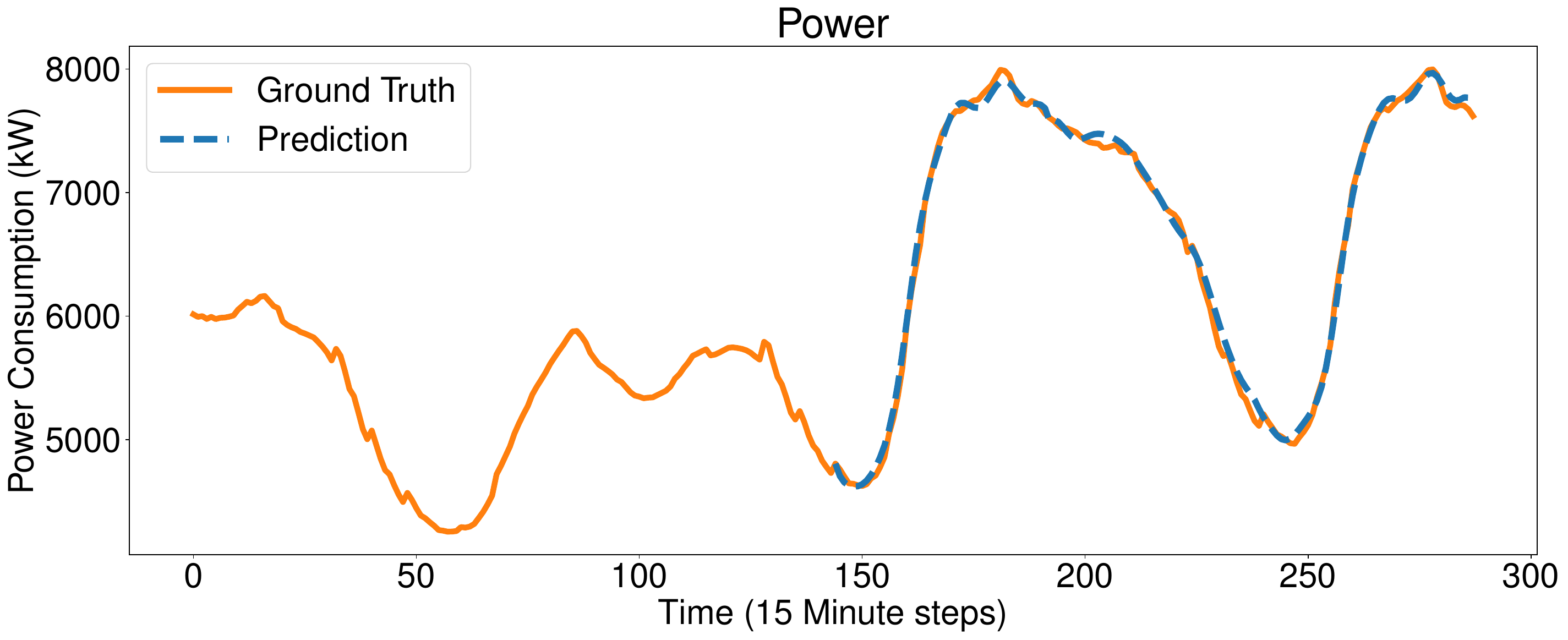}
            \end{minipage}
            \begin{minipage}[b]{.485\textwidth}
              \centering
              \includegraphics[width=0.9\textwidth]{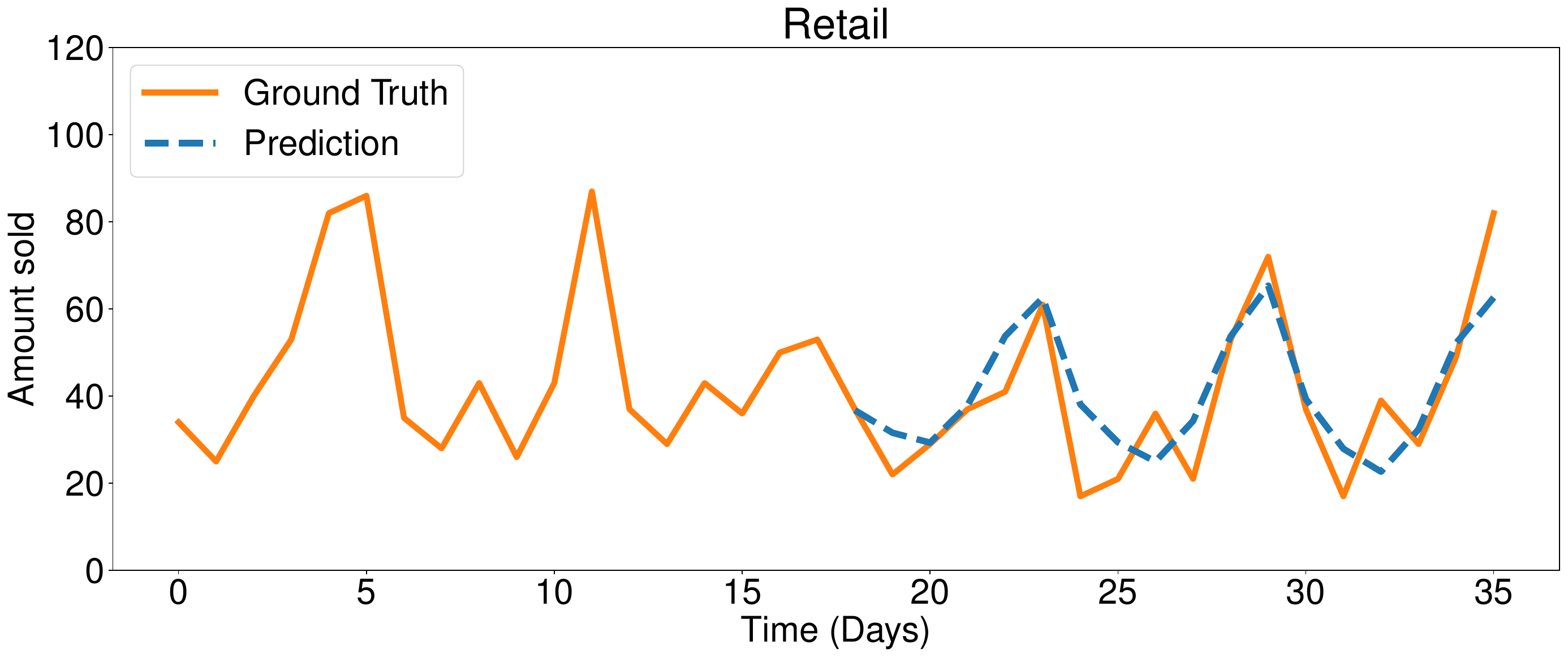}
            \end{minipage}
            \caption{Predictions on \textit{Power} (left) and \textit{Retail} dataset (right). Context is not shown for clarity.
            We can observe much bigger variance between time steps for \textit{Retail} data. This makes the prediction task for the SRNN harder.}
            \label{fig:DataExample}
        \end{figure*}
        
        We alter this approach to account for the complex values of the Fourier coefficients similarly to the Complex-Valued SPNs (CoSPNs)~\citep{yu2021icml_wspn}, resulting in Complex Conditional SPNs (CoCSPNs).
        The input for the leaves of the CoCSPN are the Fourier coefficients of $Y$ at frequency $k$ and shift $m$, i.e. $d_k^m = STFT(Y)_k^m$.
        Based on the Whittle assumption, we know that the Fourier coefficients $d_k^m$ are normal distributed.
        Therefore, also their real and imaginary parts are jointly normal distributed.
        To account for the correlations between the two parts, they are jointly modeled by a single pairwise Gaussian leaf node, parameterized by a vector of means $\mu_{d_k^m} \in \mathbb{R}^2$ and a covariance matrix $\Sigma_{d_k^m} \in \mathbb{R}^{2 \times 2}$.
        Thus, CoCSPN encodes the conditional
        \begin{equation}
                p(d_1^1, \ldots, d_{\mathcal{T}}^1, \ldots, d_1^{n_s}, \ldots, d_{\mathcal{T}}^{n_s} | STFT(X)),
        \end{equation}
        where $d_k^m = [\operatorname{Re}(d_k^m), \operatorname{Im}(d_k^m)]$.
        Based on this and Equation \ref{eq:whitle_likelihood_0}, we define the CWLL as: 
        \begin{equation}
    	    \begin{aligned}
        		\ell(&d_1^1, \ldots, d_{\mathcal{T}}^1, \ldots, d_1^{n_s}, \ldots, d_{\mathcal{T}}^{n_s} | STFT(X))   \\
        		&=\log p(d_1^1, \ldots, d_{\mathcal{T}}^1, \ldots, d_1^{n_s}, \ldots, d_{\mathcal{T}}^{n_s} | STFT(X)),
    		\end{aligned}
    	\end{equation}
        which models the probability of the predicted STFT-windows given the STFT-windows of the context, i.e. the Fourier coefficients of $STFT(X)$ are provided to the neural network $g$.
        The structural constraints of completeness and decomposability for SPNs still hold, as in~\citep{yu2021icml_wspn}.
        
    \subsection{Spectral RNN}
    \label{sec:SRNN}
    
        We employ \nameref{sec:SRNN}~\citep{wolter2020sequence}, designed for univariate time series forecasting, to provide predictions in RECOWN.
        Compared to a standard RNN, the recurrent steps are performed over the windows retrieved from STFT.
        Therefore, for a window with width $T_w$ and a step size $S$, it only has to perform $n_s$ instead of $T$ time steps for an input of length $T$.
        The SRNN is defined as follows:
        \begin{equation}
            \mathbf{X}_{\tau}= STFT(X)^\tau
        \end{equation}
        \begin{equation}
            \mathbf{z}_{\tau}=\mathbf{W}_{c} \mathbf{h}_{\tau-1}+\mathbf{V}_{c} \mathbf{X}_{\tau}+\mathbf{b}_{c}
        \end{equation}
        \begin{equation}
            \mathbf{h}_{\tau}=f_{a}\left(\mathbf{z}_{\tau}\right)
        \end{equation}
        \begin{equation}
            \mathbf{y}_{\tau}=iSTFT(\mathbf{W}_{p c} \mathbf{h}_{0}, \ldots, \mathbf{W}_{p c} \mathbf{h}_{\tau}),
        \end{equation}
        with $\tau=\left[0, n_{s}\right]$ enumerating the total number of segments $n_{s}$.
        Since $\mathbf{X}_{\tau} \in \mathbb{C}^{\mathcal{T}} \times 1$ is a complex signal, the RNN cell either needs to operate in the complex space or needs to provide projections $\mathcal{I}: \mathbb{C}^{\mathcal{T}} \mapsto \mathbb{R}^{n_i}$, $\mathcal{O}: \mathbb{R}^{n_o} \mapsto \mathbb{C}^{\mathcal{T}}$ for $n_i$-dimensional in- and $n_o$-dimensional outputs respectively.
        Since complex units have not shown superior in preliminary experiments (similar to \citep{wolter2020sequence}), we employ standard Gated Recurrent Units (GRU)~\citep{chung2014empirical}.
        As projections, we employ concatenation and splitting respectively, i.e. $\mathcal{I}(X_{\tau}) = (Re(X_{\tau}), Im(X_{\tau}))$, $\mathcal{O}(h_{\tau}) = h_{\tau}^{1, ..., \mathcal{T}} + h_{\tau}^{\mathcal{T} + 1, ..., 2 \mathcal{T}} \cdot i$ and $n_i = n_o = 2 \mathcal{T}$.
        Thus, $\mathbf{h}_{\tau} \in \mathbb{R}^{n_{h} \times 1}, \mathbf{W}_{c} \in \mathbb{R}^{n_{h} \times n_{h}}, \mathbf{V}_{c} \in \mathbb{R}^{n_{h} \times 2 \mathcal{T}}, \mathbf{b}_{c} \in \mathbb{R}^{n_{h} \times 1}$ and $\mathbf{W}_{p c} \in \mathbb{R}^{n_{h} \times 2 \mathcal{T}}$, where $n_{h}$ is the size of the hidden state and $\mathcal{T}$ the reduced amount of frequencies in the STFT.
        Additional details on SRNN are in \citep{wolter2020sequence}.
    
    \subsection{Training RECOWNs}
    
        \begin{figure*}[t!]
            \graphicspath{{./plots/}}
            \centering
            \begin{minipage}[b]{.49\textwidth}
              \centering
              \includegraphics[width=0.9\textwidth]{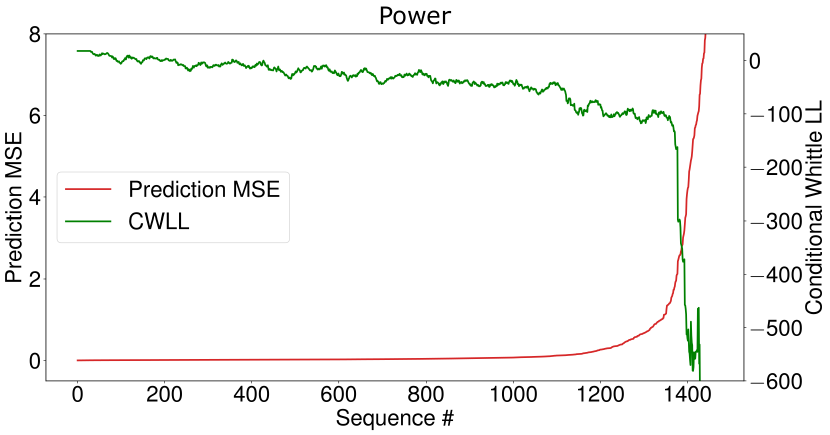}
            \end{minipage}
            \begin{minipage}[b]{.49\textwidth}
              \centering
              \includegraphics[width=0.9\textwidth]{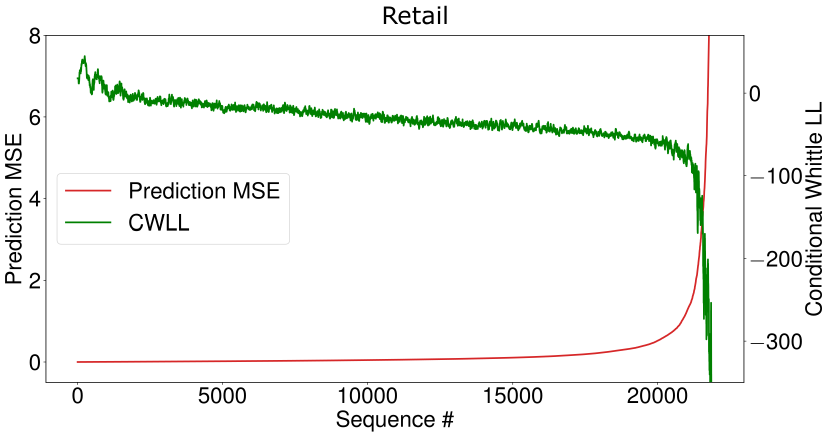}
            \end{minipage}
            \caption{RECOWNs can correctly separate between bad and good predictions. This is captured by the correlation of CWLL and MSE on \textit{Power} (left) and \textit{Retail} dataset (right). On the x-axis is denoted the enumeration of all test sequences (both context and prediction) in ascending order by MSE. It can be observed a clear (negative) correlation between a decreasing CWLL and an increasing MSE. The CWLL is smoothed by a moving average of 12 for clarity.}
            \label{fig:Separation}
        \end{figure*}
        
        \begin{figure}[t!]
            \graphicspath{{./plots/}}
              \centering
              \includegraphics[width=0.9\linewidth]{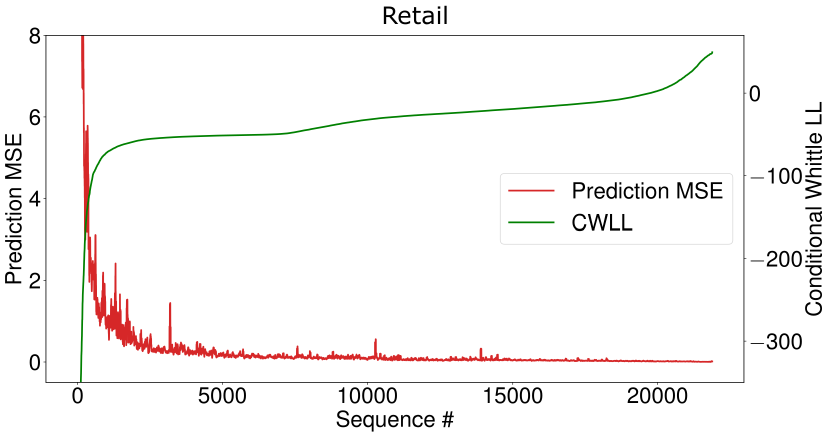}
            \caption{RECOWNs CWLL can be employed successfully to sort sequences w.r.t. their expected error when the ground truth is not available. This represents a real use case, where possibly ``bad'' forecasts need to be detected.. 
            For clarity, the MSE is smoothed by a moving average of 12.}
            \label{fig:OrderByLL}
        \end{figure}
        
        The architecture of RECOWN is shown in Figure~\ref{fig:SystemStructure}.
        SRNN and CWSPN can be trained end-to-end in a co-ordinate descent fashion, as shown in Figure~\ref{fig:Training}, resulting in RECOWN. 
        In each optimization step, the weights of the SRNN are updated first by minimizing the Mean Squared Error (MSE-loss), i.e. $MSE(Y_{GT}, Y_{Pred})$.
        Afterwards, the SRNN weights are fixed and the CWSPN is to be optimized w.r.t. minimizing an MSE-weighted negative Log-Likelihood (LL) loss on $M$ samples of data:
        \begin{equation}
        \label{eq:MSEWLLLoss}
            -\frac{1}{M} \sum_{i=0}^{M} \frac{\ell(Y_{Pred}^i | X^i)}{SE(Y_{GT}^i, Y_{Pred}^i)^{2} },
        \end{equation}
        where $SE$ denotes the Squared Error.
        Therefore, the CWSPN learns a skewed conditional distribution of predictions given the context: Predictions of the SRNN are directly passed to the SPN-part of CWSPN, while the Fourier coefficients of the context are passed to the NN $g$ of CWSPN.
        By using the inverse weighting with the SE, we assume that less accurate predictions (high SE) occur less frequently in the distribution, while better predictions (low SE) are assumed to appear more often in the distribution.
        We opt for the square of the scaling factor to account for the exponentially-shaped curve of the MSE as it can be seen in Figure \ref{eq:FT}.
        This helps the CWSPN to learn a more meaningful likelihood, as we will show in Section \ref{sec:Experiments}.
        One can employ any gradient-descent-based density estimator in RECOWN. However,
        as shown in the next Section, CWSPNs turned to be more suitable for RECOWN.
    
\section{Experimental Evaluation}
\label{sec:Experiments}

    To show the benefits of using RECOWNs, we investigate the following research questions:
    \begin{description}
        \item[(Q1) Quality of Predictions] Can RECOWNs distinguish between ``bad'' and ``good'' predictions on a sequence level?
        \item[(Q2) Uncertainty of Predictions] Using the Log-Likelihood Ratio Score (proposed in Section \ref{sec:Uncertainty}), can RECOWNs estimate the uncertainty of predictions in the time domain?
    \end{description}

    \subsection{Datasets}

        For our experiments\footnote{Code at: \url{https://github.com/ml-research/wtf-nets}} we evaluate the performance on two different
        z-score normalized datasets. 

        The first dataset is \textit{Power} consumption, which uses the 15 minute-frequency data from the European Network of Transmission System Operators for Electricity~\footnote{\url{https://transparency.entsoe.eu/} - we use the version crawled and made available by~\citet{wolter2020sequence}}. 
        Given 14 days of context, the network has to predict the power load from noon to midnight of the following day (i.e. 1.5 days).
        As window size, we choose 96, which corresponds to a full day given the 15-minute sampling rate.
        
        The second dataset regards the task of forecasting the \textit{Retail} demand, using data from a retail location of a big German retailer~\footnote{We cannot unveil the name of the company due to an NDA.}, spanning over 2 years and including roughly 4k different products with a daily sampling rate.
        Here, the task is to predict three weeks of demand given half a year of context.
        Since no sales data of Sundays is present, we filter them out, making a window size of 12 a reasonable choice, i.e. spanning 2 weeks.
        We deliberately use a smaller window size here compared to the \textit{Power} dataset in order to verify that the approach works also with different window sizes.
        Regarding the low-pass filter of STFT, we apply it with a factor of $4$ to the \textit{Power} and with a factor of $2$ to the \textit{Retail} data. 
        Examples of sequences of both datasets can be found in Figure \ref{fig:DataExample}, together with the predictions of the SRNN.

    \subsection{Classification Accuracy} 
    \label{sec:PredQuality}

        Since the target of our approach is to use the likelihood of the CWSPN as an indicator for the quality of predictions, we will focus our evaluation on whether this correlation is true for a significant part of our test sequences or not.
        To do so, we first inspect the correlation visually through plots and revise it more formally later by using a correlation error.
        
        \begin{table*}[t!]
            \begin{center}
                \small
                \begin{tabular}{c|rrr|rrr|}
                    \cline{2-7}
                    & \multicolumn{6}{c|}{\textit{Test Correlation Error}} \\
                    \cline{2-7} 
                     & \multicolumn{3}{c|}{\textit{Power}} & \multicolumn{3}{c|}{\textit{Retail}} \\
                    \cline{2-7} 
                     &  Small & Medium & Large & Small & Medium & Large \\
                    \hline
                    \multicolumn{1}{|c|}{\textit{RECOWN}}  & \textbf{0.019} & \textbf{0.016} & \textbf{0.011} & \textbf{0.036} & 0.035 & 0.027 \\
                    \hline 
                    \multicolumn{1}{|c|}{\textit{RECOWN (Time)}}  & 0.023 & 0.019 & 0.017 & 0.042 & \textbf{0.031} & 0.030 \\
                    \hline 
                    \multicolumn{1}{|c|}{\textit{RECOWN-MAF}}  & 0.045 & 0.026 & \textbf{0.011} & 0.044 & 0.033 & \textbf{0.023} \\
                    \hline 
                    \multicolumn{1}{|c|}{\textit{RECOWN-MAF (Time)}}  & 0.093 & 0.058 & 0.051 & 0.047 & 0.045 & 0.029 \\
                    \hline 
                    \multicolumn{1}{|c|}{\textit{Random}}  & \multicolumn{3}{c|}{0.400} & \multicolumn{3}{c|}{0.455} \\
                    \hline
                    \hline
                    \multicolumn{1}{|c|}{\#Parameters} & 300k & 900k & 3M & 30K & 70K & 200K \\
                    \hline 
                    \hline
                \end{tabular}
            \end{center}
            \caption{Test Correlation Error (lower is better) for different architectures modeling the time
            series in the time domain (denoted with ``Time'') or in the spectral domain. 
            Lower score indicates a stronger correlation between CWLL and MSE.
            The results indicate that RECOWNs are trustworthy models which can distinguish between good and bad predictions. Besides, modeling in the spectral domain generally outperforms modeling in the time domain w.r.t. the Correlation Error, in particular for MAF, where it considerably improves parameter efficiency.
            Furthermore, for smaller model sizes, RECOWNs achieve the best scores, while MAF is better for models with larger capacity. }
            \label{tbl:CE}
        \end{table*}
        
        \subsubsection{Correlation Plots}
        
            To visually inspect the quality of the CWLL as an indicator for the prediction quality, we first calculate the CWLL and MSE for all test sequences of each dataset and plot those in ascending order by MSE.
            As shown in Figure~\ref{fig:Separation}, a clear correlation between CWLL and MSE exists: The higher the MSE, the lower the CWLL.
            The same happens for the magnitude, i.e. the exponentially-shaped curve of the MSE is reflected in the CWLL.
            Based on these observations, we can assume that the CWLL is a good indicator for the prediction quality, and it fulfills \textbf{(Q1)}.
            We show this assumption holds more formally in Section \ref{sec:CE}.

            As in real-world scenarios, where usually the ground truth is not (yet) known and the MSE is therefore missing, one could use the CWLL to sort sequences from potential high MSE (low CWLL) to probably low MSE (high CWLL) and analyze further those potential high MSE frequencies.
            Such a use case is depicted in Figure~\ref{fig:OrderByLL}. It shows once again a strong correlation between an increasing CWLL with a lower MSE of predictions.
            To have a quantitative perspective, we provide the following example of a potential use-case: Selecting the 5\% sequences with the lowest CWLL from CWSPN on \textit{Power}, we find that 75\% of all sequences with the worst 5\% of MSEs are included. 
            Looking at the 10\% sequences with the lowest CWLL, we find 98.5\% of all sequences with the worst 5\% of MSEs.
            This highlights how RECOWNs are able to distinguish between good and bad predictions.
        
        \subsubsection{Correlation Error (CE)}
        \label{sec:CE}
 
            In order to provide a correlation error for each test sequence $n$, we calculate a relative prediction error $S_{Pred}^n =$
            \begin{equation}
                \sqrt{\frac{SE(Y_{Pred}^n, Y_{GT}^n) - \min_m SE(Y_{Pred}^m, Y_{GT}^m)}{\max_m SE(Y_{Pred}^m, Y_{GT}^m) - \min_m SE(Y_{Pred}^m, Y_{GT}^m)}},
            \end{equation}
            and a likelihood score ($S_\ell^n$) respectively:
            \begin{equation}
                S_\ell^n = \sqrt{\frac{\ell(Y_{Pred}^n | X^n) - \max_m \ell(Y_{Pred}^m | X^m)}{\min_m \ell(Y_{Pred}^m | X^m) - \max_m \ell(Y_{Pred}^m | X^m)}}.
            \end{equation}
             The square root is taken to take into account the exponential shape of the conditional CWLL, as can be seen in Figure~\ref{fig:Separation}.
            
            Given that the MSE reflects the gold standard on where a sequence should be placed on the range from ``bad'' to ``good'', we define the Correlation Error (CE) for the CWLL as the quadratic distance of the scores:
            \begin{equation}
                CE^n = (S_{Pred}^n - S_\ell^n)^2.
            \end{equation}
            Since $S_{Pred}^n, S_\ell^n \in [0, 1]$ by definition, we have $CE^n \in [0, 1]$.
            In order to better assess this novel score, we provide a random baseline, which draws likelihood scores randomly from a uniform distribution, i.e. $S_{\ell_{random}}^n \sim \mathbf{U}(0, 1)$.

            To evaluate the correlation error, we compare CWSPN with Masked Autoregressive Flow (MAF)~\citep{papamakarios2017masked}, a state-of-the-art neural density estimator.
            MAF is integrated into the joint RECOWN architecture like CWSPN, therefore, it follows the same training objective as given in Equation~\ref{eq:MSEWLLLoss}.
            We refer to this architecture as \textit{RECOWN-MAF}.
            For each model, we report scores from the spectral domain as well as on the original time series (for CWSPN, modeling the time series in the time domain results in a CSPN).
            Furthermore, we evaluate three different model sizes, \textit{Small, Medium,} and \textit{Large}.
            The results and the number of trainable parameters are given in Table~\ref{tbl:CE}.

            In general, modeling in the spectral domain (by using STFT) is more beneficial than operating in the time domain, while improving also parameter efficiency.
            This is more prominent for MAF.
            Furthermore, MAF achieves the best scores on bigger models, in comparison, CWSPN is particularly good with reduced model capacity.
            Overall, the correlation error obtained with the different architectures is relatively low, also on \textit{Retail} which is a more difficult dataset.
            Moreover, it is much better than the random baseline.
            This allows us to answer \textbf{(Q1)} affirmatively.
            -- RECOWN can distinguish between ``bad'' from ``good'' predictions on a sequence level. 
            However, it is important to remark that SPN architectures can naturally answer a wider range of queries than MAF.
            Additionally, during our experiments, we observed that CWSPN is also less sensitive to hyperparameter tuning.
    
    \subsection{Providing Uncertainty to Predictions}
    \label{sec:Uncertainty}

        The conditional Whittle Likelihood can also be used to estimate the uncertainty for a prediction in the time domain.
        This allows to take insights on how it changes over time.
        With this aim, we use the notion of likelihood ratios, leveraging the window function $w$, to project the likelihood to the time domain at time step $n$:
        \begin{equation}
    		\lambda_{LR}(n) = -2 (w(n)\ell(Y_{pred}(n) | X) - \max_{Y, X \in D_{\textrm{Train}}} \ell(Y | X)).
    	\end{equation}
    	Here, $\ell(Y_{pred}(n) | X)$ denotes the conditional likelihood of prediction window $n$ given the context.
    	Every other window in the prediction is marginalized.
    	Note that the calculation of $\lambda_{LR}$ has high similarities to the likelihood-score $S^n_{\ell}$ introduced in Section \ref{sec:CE}.
    	Furthermore, we define the maximum likelihood ratio occurring in the training data:
    	\begin{equation}
        		\lambda_{LRmax} = -2 \left( \; \min_{Y, X \in D_{\textrm{Train}}} \ell(Y | X) -\max_{Y, X \in D_{\textrm{Train}}} \ell(Y | X) \right).
    	\end{equation}
    	Using  $\lambda_{LRmax}$ as normalization for $\lambda_{LR}$, we can estimate the uncertainty of the prediction via the Log-Likelihood Ratio Score (LLRS):
    	\begin{equation}
    		LLRS(n) = \sqrt{\frac{\lambda_{LR}(n)}{\lambda_{LRmax}}}.
    	\end{equation}
        As with the correlation error, we take the square root to account for the exponential shape of the CWLL (see e.g. Figure~\ref{fig:Separation}).
        In this way, a likelihood equally worse as the worst training sample likelihood (i.e. $\ell(Y_{pred} | X) = \min_{Y, X \in D_{\textrm{Train}}} \ell(Y | X)$) results in $LLRS = 1$.
        Larger likelihoods (i.e. $\ell(Y_{pred} | X) > \min_{Y, X \in D_{\textrm{Train}}} \ell(Y | X)$) result in scores $LLRS < 1$, smaller likelihoods (i.e. $\ell(Y_{pred} | X) < \min_{Y, X \in D_{\textrm{Train}}} \ell(Y | X)$) in $LLRS > 1$.
        Thanks to z-score normalization, LLRS can be applied without concerns about the magnitude of the original data. 
        
        To evaluate LLRS potential, we run long-time range forecasting on both datasets.
        For the \textit{Power} dataset, we predict 40 days as long-range prediction, performed by RECOWN trained only for 5 days prediction (``standard prediction'').
        For the \textit{Retail} dataset, we predict 16 weeks, performed by RECOWN trained only for 8 weeks prediction.
        Our assumption is that the predictions made by the SRNN are less accurate over time and our aim is to make sure that 
        this undesirable behaviour is captured by the uncertainty estimated with LLRS.
        As it can be seen in Figure~\ref{fig:Uncertainty}, the more the prediction diverges from the ground truth over time, the more the uncertainty grows.
        This shows that the uncertainty estimated by LLRS can be a good indicator for users to detect such cases, even when the ground truth might be not available.
        Therefore, \textbf{(Q2)} can be answered positively:  
        RECOWNs can estimate the uncertainty of predictions in the time domain and this can support users in decision-making.
    
\section{Conclusion}
\label{sec:Conclusion}

    In this paper, we proposed RECOWNs, a novel architecture that jointly trains an SRNN and a probabilistic circuit in order to provide the SRNN with a measure of uncertainty of its predictions, on sequence level as well as in the time domain.
    This yields a trustworthy model for time series forecasting being able to inform the users when it is not confident of its predictions. 
    We leveraged the conditional likelihood of context and predictions together with the Whittle approximation to introduce CWSPNs, which can provide a likelihood of the prediction given the context in the spectral domain. 
    Furthermore, we introduced LLRS, an effective score to evaluate the uncertainty for any time point of a prediction. 
    Our experiments on real-world datasets show that RECOWNs are both accurate and trustworthy. 
    In this context, a probabilistic circuit tailored for time series based on the Whittle approximation showed to be superior to MAF, a state-of-the-art density estimator.
    We hope our results will inspire further research on trustworthy models for time series forecasting.
    
    Future work may extend our contributions in several ways.
    Since RECOWN can be trained in an end-to-end fashion, we envision more sophisticated strategies for joint training.
    For example, gradients from the CWSPN could be used to improve the predictions of an SRNN.
    Moreover, it could be explored whether RECOWNs are robust against adversarial attacks.
    Besides, tractable general inference of CWSPNs could be exploited to gain knowledge about factors of influence for the prediction.
    In the same direction, modeling the joint distribution of the coefficients instead of the conditional, thus,
    being able to compute any marginal, could open new opportunities. 
    For this scope, employing Einsum Networks \citep{peharz2020einsum},
    fast and scalable probabilistic circuits, could be an effective solution.

\begin{acknowledgements}
This work was supported by the Federal Ministry of Education and Research (BMBF; project ``MADESI'', FKZ 01IS18043B, and Competence Center for AI and Labour; ``kompAKI'', FKZ 02L19C150), the German Science Foundation (DFG, German
Research Foundation; GRK 1994/1 ``AIPHES''), the Hessian Ministry of Higher Education, Research, Science and the Arts (HMWK;
projects ``The Third Wave of AI'' and ``The Adaptive Mind''), and the
Hessian research priority programme LOEWE within the
project ``WhiteBox''.
The authors thank German Management Consulting GmbH for supporting this work.
\end{acknowledgements}

\bibliography{uai2021-template}

\end{document}